\documentclass[journal]{IEEEtran}
\usepackage{booktabs}
\usepackage{tabularx}
\usepackage{cite}
\usepackage{amsmath}
\usepackage{amsfonts}
\usepackage{graphicx} 
\usepackage{csvsimple,booktabs}
\usepackage{float,enumitem}
\usepackage{graphicx,epsfig,xcolor}
\usepackage{balance}
\usepackage[hyphens]{url}
\usepackage[breaklinks, colorlinks, linkcolor=purple, anchorcolor=purple, citecolor=purple, urlcolor=purple]{hyperref}
\usepackage{amsmath}
\usepackage{algorithm}
\usepackage{algpseudocode}
\usepackage{romannum}
\newcommand{\RNum}[1]{\uppercase\expandafter{\romannumeral #1\relax}}
\usepackage{subcaption}
\usepackage{amsthm}

\theoremstyle{definition}   

\usepackage{mathtools}
\usepackage[table]{xcolor}
\usepackage{adjustbox}
\usepackage{empheq}
\usepackage{xcolor}
\usepackage[normalem]{ulem}
\usepackage{multirow} 

\newcommand{\best}[1]{\textcolor{red}{\textbf{#1}}}
\newcommand{\second}[1]{\textcolor{blue}{\underline{#1}}}

\ifCLASSINFOpdf
\else
\fi
\begin{document}

%
\title{A Behavior-Guided Online Probabilistic Forecasting Method for Electric vehicle Charging Loads}
%
%
%

\author{
Chenghan~Li,~\IEEEmembership{Graduate Student Member,~IEEE,}
Qingxiang~Liu,~\IEEEmembership{Graduate Student Member,~IEEE,}
Yinliang~Xu,~\IEEEmembership{Senior Member,~IEEE,}
and~Yuxuan~Liang,~\IEEEmembership{Member,~IEEE}%
\thanks{Chenghan~Li and Yinliang~Xu are Tsinghua Shenzhen International Graduate School, Tsinghua University, Shenzhen 518055, China
(Corresponding author e-mail: xu.yinliang@sz.tsinghua.edu.cn)}
\thanks{Qingxiang Liu and Yuxuan Liang is with the Intelligent Transportation Thrust,
The Hong Kong University of Science and Technology (Guangzhou),
Guangzhou 511458, China.}%
}

\maketitle

\begin{abstract}
Electric vehicle (EV) charging loads exhibit strong behavioral 
heterogeneity and temporal variability, posing significant challenges for online probabilistic forecasting under evolving operating conditions. In particular, persistent charging patterns may differ substantially across stations, while recent behavioral changes can continuously alter the underlying load distributions. This paper proposes a behavior-guided online probabilistic forecasting framework that explicitly characterizes persistent station-specific patterns and recent behavioral changes. A dual-timescale behavior representation is constructed to distinguish long-term charging characteristics from recent behavioral states and quantify their deviations. These behavioral changes are further semantically encoded to guide drift-aware forecasting adaptation, while a delayed-feedback mechanism ensures temporally consistent online updates 
when observations become available across different forecasting horizons. Experiments on ten heterogeneous real-world charging stations demonstrate that the proposed method consistently outperforms conventional forecasting models and concept-drift-aware online baselines in forecasting accuracy and probabilistic reliability. For 1-h-ahead forecasting, the proposed method reduces mean squared error(MSE) and Pinball loss by 15.3\% and 17.8\%, respectively, over the corresponding best baselines. For 4-h-ahead forecasting, the improvements further reach 16.8\% and 22.6\%, respectively, demonstrating consistent performance gains under evolving charging behaviors and extended forecasting horizons.
\end{abstract}

\begin{IEEEkeywords}
EV charging load forecasting, online forecasting, charging behavior, concept drift.
\end{IEEEkeywords}

%
\IEEEpeerreviewmaketitle

\section{Introduction}
The rapid penetration of EVs is reshaping electricity demand and introducing increasingly flexible yet uncertain loads into power systems~\cite{shariatzadeh2025charging,wang2023heterogeneous}. Accurate forecasting of EV charging loads is therefore essential for charging infrastructure operation, demand-side management, and reliable 
gridoperation~\cite{zhang2021probabilistic,chemudupaty2025uncertain}. Unlike conventional loads, EV charging demand is strongly driven by individual mobility and charging decisions, resulting in substantial heterogeneity in charging time, duration, energy demand, and charging frequency across users and stations~\cite{kreft2024predictability,muller2024household}. Such behavioral heterogeneity can propagate to the grid level, producing markedly different temporal and spatial load 
profiles and increasing system-level load variability
~\cite{behavioral2026uncertainty}\cite{dimitriadou2024forecasting}\cite{behavioral2025infrastructure}. Consequently, future charging loads may deviate substantially from historically learned patterns, making them difficult to characterize using deterministic forecasts alone. Probabilistic forecasting, which represents predictive uncertainty through conditional distributions, quantiles, or prediction intervals, therefore provides a more informative characterization of future charging 
demand and is increasingly important for reliable EV charging management~\cite{li2024diffplf}.

Existing studies have extensively investigated EV charging load forecasting from temporal, spatial, and probabilistic perspectives. Wu et al.~\cite{wu2024vmd} proposed a VMD--SSA--SVR framework that decomposes historical charging loads and incorporates real-time electricity prices and ambient temperature to improve short-term forecasting accuracy. However, its decomposition and regression pipeline relies primarily on predefined numerical features and does not explicitly characterize heterogeneous charging behaviors. Wang et al.~\cite{wang2023shortterm} developed an LSTM-based model using large-scale EV trajectory data to capture temporal dependencies in station-level charging demand. Although effective for short-term prediction, the model mainly learns historical temporal regularities and provides limited capability to quantify predictive uncertainty. Cao et al.~\cite{cao2024feature} proposed a feature-enhanced deep learning framework that incorporates correlations between charging demand and external factors into probabilistic forecasting. However, the learned feature relationships are mainly derived from historical data and do not explicitly describe how station-specific charging behaviors evolve over time.

Beyond temporal and probabilistic modeling, recent studies have increasingly explored richer spatio-temporal representations and heterogeneous information for EV charging forecasting. Li et al.~\cite{li2024diffplf} introduced a conditional diffusion model to approximate the predictive distribution of EV charging loads and generate reliable probabilistic forecasts under stochastic charging demand. Nevertheless, its focus is primarily on modeling predictive distributions rather than adapting the forecasting process to continuously evolving behavioral patterns. Dai and Shi~\cite{dai2026multisource} developed a multi-source spatio-temporal framework that integrates heterogeneous station attributes, historical observations, and inter-station dependencies for fine-grained charging demand forecasting. However, the model is primarily designed to learn spatio-temporal relationships from predefined inputs, while temporal changes in station-specific behavior are not explicitly modeled. More recently, Tu et al.~\cite{tu2026kastar} proposed a Kolmogorov--Arnold spatio-temporal attention recurrent network to jointly capture spatial dependencies and nonlinear temporal patterns for multi-target EV charging load forecasting. Despite its enhanced spatio-temporal representation, the forecasting mechanism still mainly relies on learned historical dependencies and does not explicitly distinguish persistent charging patterns from recent behavioral deviations. Overall, these studies have substantially advanced EV charging forecasting, but most focus on improving temporal, spatial, or uncertainty representations under historically learned patterns.

To further address the evolving nature of charging demand, recent studies have increasingly investigated online forecasting to accommodate non-stationarity and concept drift.
Pham et al.~\cite{pham2023fsnet} proposed FSNet, which combines fast adaptation with associative memory to respond to new patterns while retaining recurring knowledge. However, its adaptation is driven primarily by changes in temporal representations rather than explicit behavioral information. Zhang et al.~\cite{zhang2023onenet} introduced OneNet to dynamically ensemble complementary forecasters for improved 
adaptation to concept drift, but the detected distribution changes remain difficult to associate with interpretable behavioral variations. Lau et al.~\cite{lau2025dsof} developed a dual-stream framework that separates fast adaptation from slow learning and avoids information leakage in online forecasting. Nevertheless, its adaptation mainly focuses on balancing recent and historical temporal information. Cai et al.~\cite{cai2025lstd} disentangled long- and short-term latent 
states to improve adaptation under unknown interventions; however, these states are learned implicitly and do not explicitly characterize domain-specific behavioral changes. Zhao and Shen~\cite{zhao2025proceed} proposed PROCEED to estimate concept drift and proactively translate it 
into model adaptation, while the drift is characterized mainly from statistical changes between training and test samples. More recently, Chen et al.~\cite{chen2025leaf} distinguished macro- and micro-drift and introduced a meta-learning framework to accommodate both long-term and short-term distribution changes. Despite these advances, existing online forecasting methods predominantly characterize non-stationarity from data distributions or latent representations, leaving the behavioral mechanisms underlying EV charging-load evolution largely unexplored.

Despite these advances, a critical gap remains between EV charging behavior analysis and online load forecasting. Existing behavioral studies have demonstrated that EV users exhibit substantial heterogeneity in charging timing, duration, frequency, and energy demand\cite{kreft2024predictability,hardman2024styles}. Such 
heterogeneity is closely associated with vehicle usage patterns, charging environments, and user preferences 
\cite{zhang2026usage,lee2025public}. Moreover, recent evidence suggests that charging characteristics are not necessarily stationary, as user engagement and the effects of behavioral features may evolve over time \cite{meza2025survival}. However, these behavioral insights are typically used for offline user segmentation, charging-pattern interpretation, or static feature construction, rather than being explicitly incorporated into the online forecasting process. Conversely, existing concept-drift-aware forecasting methods can adapt model 
parameters to distributional changes, but generally represent such changes through prediction errors, latent states, or statistical distribution shifts, without describing \emph{what charging behavior has changed} and \emph{how it deviates from the station's persistent behavioral pattern}. Consequently, there remains a lack of a unified forecasting framework that explicitly distinguishes persistent station-specific charging behavior from recent behavioral evolution and translates their deviation into actionable information for online probabilistic adaptation. Bridging this gap is essential for maintaining accurate and reliable forecasts under heterogeneous and continuously evolving EV charging behaviors.

To address the above research gaps, our core contribution lies in proposing a behavior-guided online probabilistic forecasting framework for EV charging loads under heterogeneous and evolving charging behaviors. Rather than treating temporal variations solely as statistical distribution shifts, the proposed framework explicitly characterizes persistent station-specific charging patterns and their recent behavioral deviations, and further translates such behavioral changes into actionable information for online adaptation. In this manner, behavior characterization and probabilistic forecasting are integrated within a unified online framework, allowing the forecasting model to continuously adapt to evolving charging patterns while maintaining reliable uncertainty estimation. The specific contributions are summarized as follows:

\begin{itemize}

\item This paper develops a dual-timescale behavior modeling method to explicitly characterize heterogeneous and evolving EV charging behaviors. Specifically, a long-term behavior anchor is constructed from historical charging observations to represent persistent station-specific patterns, while a rolling recent behavior state continuously captures short-term behavioral changes. The deviation between these two representations explicitly describes how current 
charging behavior differs from the station-specific historical 
baseline. Compared with existing concept-drift-aware online forecasting methods~\cite{pham2023fsnet,zhang2023onenet}, which primarily adapt to distribution shifts through temporal representations or model ensembling, the proposed method characterizes charging-load evolution from a behavioral perspective, thereby providing more informative guidance for station-specific online adaptation.

\item Building upon the above dual-timescale behavior representation, this paper proposes a behavior-semantic guided probabilistic forecasting method to translate the identified behavioral deviations into actionable 
information for online adaptation. Specifically, descriptor-level prompts are constructed from the long-term behavior, recent behavior, and their deviations, which are further encoded by a frozen language model to obtain semantic representations of behavioral evolution. A lightweight adapter 
maps these representations into behavior-aware residual corrections, while a drift-aware gating mechanism dynamically regulates their contribution and the online adaptation rate according to the behavioral deviation. Compared with existing EV forecasting studies that primarily incorporate behavioral or contextual information as numerical features~\cite{cao2024feature,kreft2024predictability}, the proposed method further interprets the identified behavior changes at the semantic level and directly uses them to guide online probabilistic adaptation.

\item The proposed framework is extensively evaluated on a real-world dataset containing ten heterogeneous EV charging stations. Compared with the best-performing baselines, it reduces MSE and Pinball loss by 15.3\% and 17.8\% for 1-h-ahead forecasting, and by 16.8\% and 22.6\% for 4-h-ahead forecasting, respectively, demonstrating consistent improvements across different forecasting horizons.
\end{itemize}

The remainder of this paper is organized as follows: Section II 
formulates the online probabilistic forecasting problem for EV charging loads, including the delayed-feedback mechanism and probabilistic forecasting objective. Section III details the proposed behavior-guided online forecasting framework, including dual-timescale behavior modeling, behavior-semantic prompt encoding, and behavior-guided probabilistic 
adaptation. Section IV presents case studies on ten heterogeneous EV charging stations, with numerical results covering benchmark comparisons, ablation studies,  and computational efficiency. Section V concludes the paper and discusses future work.

\section{Problem Formulation}
\begin{figure}
    \centering
    \includegraphics[width=0.99\linewidth]{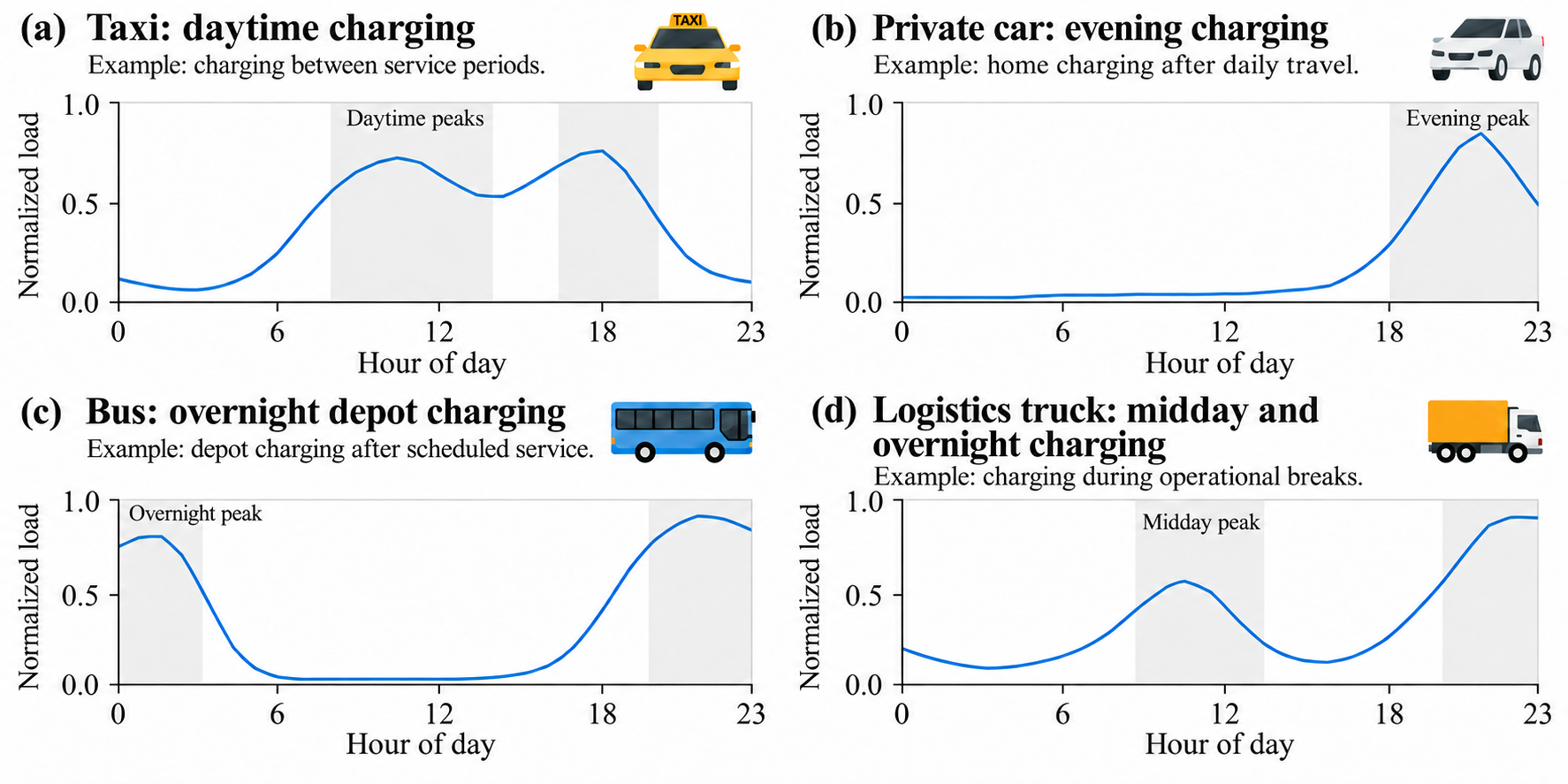}
    \caption{Heterogeneous charging behaviors across different EV types.}
    \label{fig:motivation}
\end{figure}

EV charging loads are inherently driven by users' charging behaviors, leading to heterogeneous charging patterns across stations and temporally varying behaviors within individual stations (see Figure~\ref{fig:motivation}).As these behaviors evolve over time, the underlying load patterns may deviate from historical observations, challenging forecasting models trained on fixed data. Based on this motivation, we first formulate the online EV charging load forecasting problem as follows.

Consider a set of EV charging stations
$\mathcal{S}=\{1,\ldots,S\}$. The hourly charging load of station
$s\in\mathcal{S}$ at time $t$ is denoted by
$y_{s,t}$. At each forecast origin $t$, the
historical input is given by
\begin{equation}
\mathbf{x}_{s,t}=\left[y_{s,t-L+1},\ldots,y_{s,t}\right]^{\top}\label{eq:historical_input}
\end{equation}
where $L$ is the look-back length. 

Let \(\mathcal{H}_{s,t}=\left\{y_{s,\tau}:\tau\leq t\right\}\)
denote the load history available for station $s$ at online forecast origin $t$. Given the available historical information and the calendar information associated with the target time, the $H$-step-ahead online forecasting problem can be formulated as
\begin{equation}
y_{s,t+H}
\sim
p\!\left(
y
\mid
\mathbf{x}_{s,t},
\mathbf{c}_{s,t+H},
\mathcal{H}_{s,t};
\boldsymbol{\theta}_{t}
\right)
\label{eq:online_forecasting_problem}
\end{equation}
where $\mathbf{x}_{s,t}\subset\mathcal{H}_{s,t}$ contains the most recent $L$ hourly load observations, $\mathbf{c}_{s,t+H}$ denotes the calendar covariates known for the target hour $t+H$, and $\boldsymbol{\theta}_{t}$ represents the model parameters available at forecast origin $t$.

Rather than explicitly specifying a parametric form for the conditional distribution $p(\cdot)$, we characterize the predictive distribution through a set of conditional quantiles. Specifically, the model produces
\begin{equation}
\widehat{\mathbf{q}}_{s,t+H}
=
f_{\boldsymbol{\theta}_{t}}
\left(
\mathbf{x}_{s,t},
\mathbf{c}_{s,t+H}
\right)
\label{eq:quantile_output}
\end{equation}
where \(
\widehat{\mathbf{q}}_{s,t+H}
=
\left[
\hat{q}^{\alpha}_{s,t+H}
\right]_{\alpha\in\mathcal{Q}}\)denotes the predicted quantiles at the prescribed probability levels $\mathcal{Q}$. Since the corresponding observation $y_{s,t+H}$ is not available when the forecast is issued at time $t$, the prediction cannot be immediately used for online adaptation. Instead, each forecast is retained until its target observation becomes available.
\begin{figure*}[!t]
    \centering
    \includegraphics[width=0.99\textwidth]{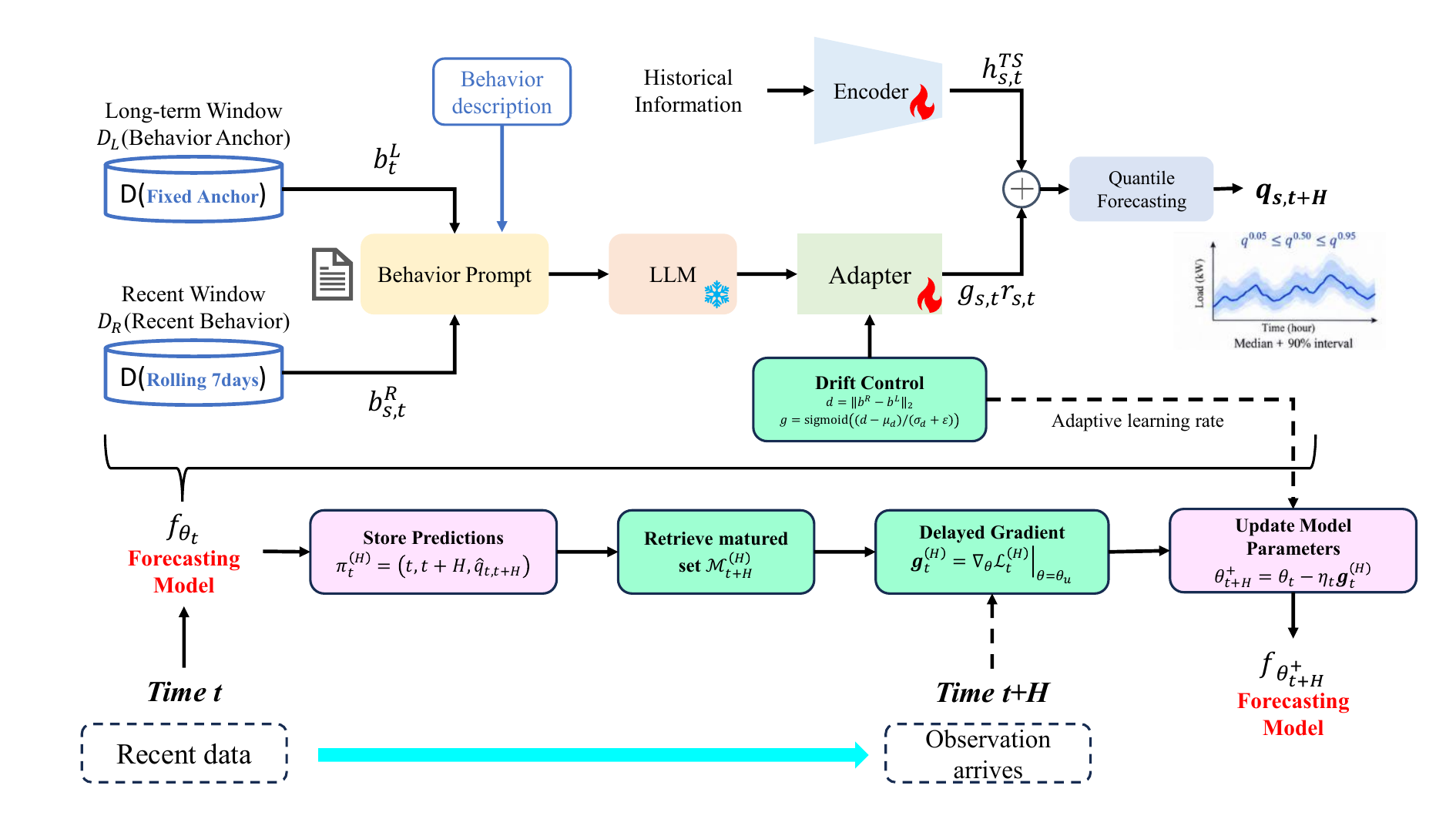}
    \caption{Framework of the proposed behavior-guided online probabilistic forecasting.}
    \label{fig:method}
\end{figure*}

Accordingly, a prediction generated at forecast origin $t$ is stored as
\begin{equation}
    \pi_{s,t}^{(H)}
    =
    \left(
        s,\,
        t,\,
        t+H,\,
        \mathbf{x}_{s,t},\,
        \mathbf{c}_{s,t+H},\,
        \widehat{\mathbf q}_{s,t+H}
    \right).
    \label{eq:pending_sample}
\end{equation}
At clock time $u$, the matured sample set is defined as
\(\mathcal{M}_{u}^{(H)}
=\left\{\pi_{s,t}^{(H)}:t+H=u,\;s\in\mathcal{S}\right\}\),
where only samples whose corresponding observations are available at time $u$ are eligible for online adaptation. The stored prediction $\widehat{\mathbf q}_{s,t+H}$ is retained for evaluating the forecast originally issued at time $t$, while $\mathbf{x}_{s,t}$ and $\mathbf{c}_{s,t+H}$ are used with the newly available observation for subsequent online adaptation.

\begin{figure}
    \centering
    \includegraphics[width=0.99\linewidth]{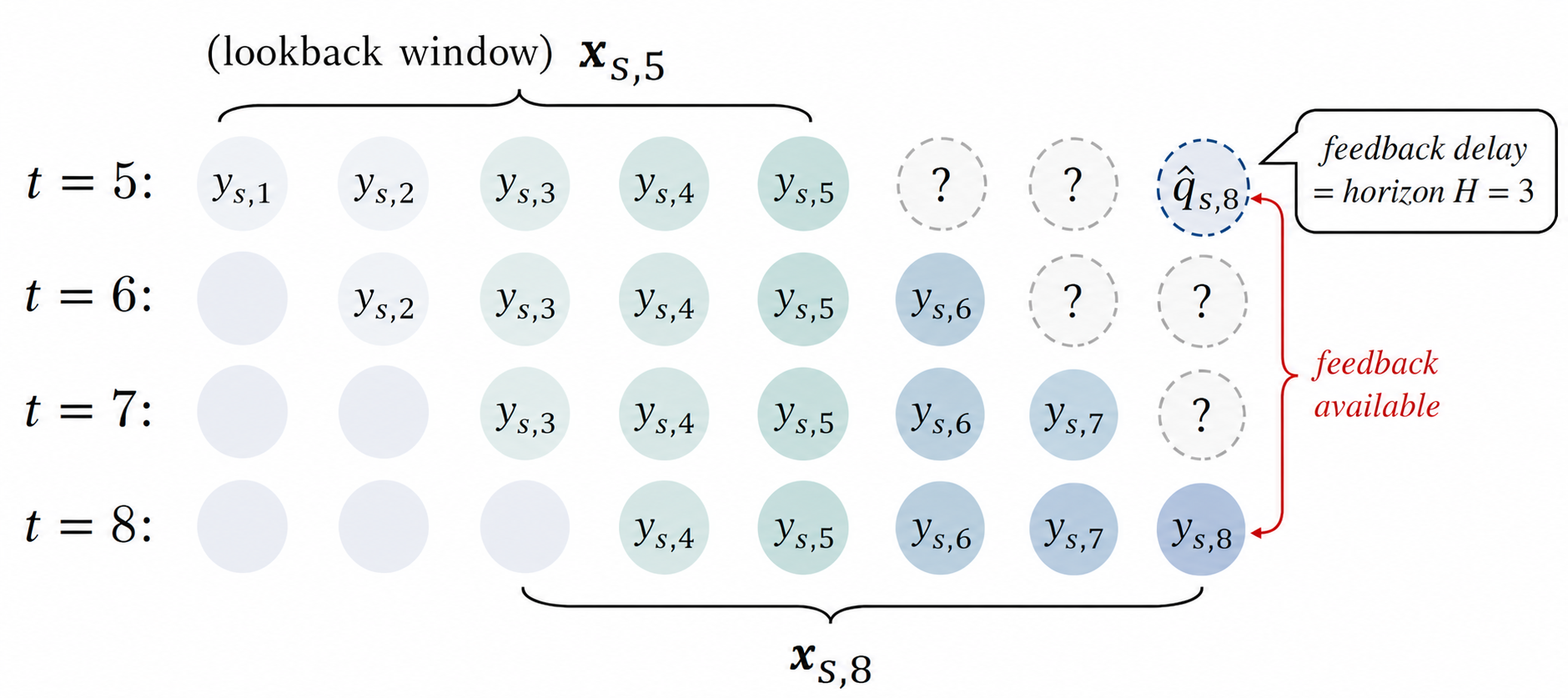}
    \caption{Delayed-feedback online forecasting with horizon $H=3$.}
    \label{fig:delay}
\end{figure}

Once a pending sample matures, the corresponding observation becomes available and its forecasting loss can be evaluated. Specifically, for a sample issued at time $t$, the gradient is computed at its target time $u=t+H$ using the current model state and the pinball loss as  
\begin{equation} 
\begin{aligned}  
&\mathbf{g}_{s,t\rightarrow u}^{(H)}=  
\left.  
\nabla_{\boldsymbol{\theta}}  
\mathcal{L}_{\mathrm{PB}}  
\left(  
\widetilde{y}_{s,u},  
f_{\boldsymbol{\theta}}  
\left(  
\mathbf{x}_{s,t},  
\mathbf{c}_{s,u}  
\right)  
\right)  
\right|_{\boldsymbol{\theta}=\boldsymbol{\theta}_{u}} 
\label{eq:delayed_gradient_a}  
\\  
&\mathcal{L}_{\mathrm{PB}}  
\left(  
\widetilde{y}_{s,u},  
\widehat{\mathbf{q}}_{s,u}  
\right)=  
\frac{1}{|\mathcal{Q}|}  
\sum_{\alpha\in\mathcal{Q}}  
\biggl[  
\alpha  
\bigl(  
\widetilde{y}_{s,u}   
\\  
&-  
\widehat{q}_{s,u}^{\alpha}  
\bigr)_{+}  
+  
(1-\alpha)  
\bigl(  
\widehat{q}_{s,u}^{\alpha}  
-  
\widetilde{y}_{s,u}  
\bigr)_{+}  
\biggr] 
\end{aligned}  
\end{equation} 
where $(z)_{+}=\max(z,0)$, $\mathcal{Q}$ denotes the set of prescribed quantile levels, and $\mathbf{g}_{s,t\rightarrow u}^{(H)}$ represents the gradient of the matured sample evaluated using the current model state. The gradient can only be evaluated when the corresponding target observation becomes available at $u=t+H$ (see Figure~\ref{fig:delay}).

Accordingly, once the gradients associated with the newly matured samples are evaluated, they are aggregated to update the current model parameters at clock time $u$ as
\begin{equation}
    \boldsymbol{\theta}_{u^{+}}
    =
    \boldsymbol{\theta}_{u}
    -
    \sum_{s\in\mathcal{S}}
    \eta_{s,u}
    \sum_{\substack{\pi_{s,t}^{(H)}\in\mathcal{M}_{u}^{(H)}}}
    \mathbf{g}_{s,t\rightarrow u}^{(H)}
    \label{eq:online_update}
\end{equation}
where $\eta_{s,u}>0$ denotes the station-specific online learning rate determined by the current behavior drift, and $\boldsymbol{\theta}_{u^{+}}$ represents the parameter state after incorporating all newly matured samples at time $u$.

Through the above delayed-feedback updates, the model sequentially minimizes the probabilistic forecasting loss over the online stream, yielding the overall objective
\begin{equation}
    \min_{\{\boldsymbol{\theta}_{t}\}_{t\geq 1}}
    \quad
    \sum_{s\in\mathcal{S}}
    \sum_{t\in\mathcal{T}_{s}}
    \mathcal{L}_{s,t}^{(H)}
    \label{eq:online_objective}
\end{equation}
where $\mathcal{T}_{s}$ denotes the sequence of online forecast origins for station $s$ and $\mathcal{L}_{s,t}^{(H)}$ is the pinball loss defined above. In addition to minimizing the forecasting loss, the resulting prediction intervals are expected to maintain their nominal coverage over the online stream. For the $90\%$ prediction interval $\widehat{\mathcal{I}}_{s,t+H}^{0.90}$, define the miscoverage indicator as $m_{s,t}^{(H)}=\mathbb{I}\{y_{s,t+H}\notin\widehat{\mathcal{I}}_{s,t+H}^{0.90}\}$. The desired long-run calibration condition is therefore
\( \left|
        \frac{1}{N}
        \sum_{t\in\mathcal{T}_{s}(N)}
        m_{s,t}^{(H)}
        -0.10
    \right|
\longrightarrow 0\). Equivalently, the empirical coverage of $\widehat{\mathcal{I}}_{s,t+H}^{0.90}$ converges to its nominal level of $0.90$.

\section{Method}
To address the above challenges, we propose a behavior-guided online forecasting framework for EV charging loads(see Figure~\ref{fig:method}). The framework integrates persistent station-specific patterns with recent behavioral changes. Specifically, historical charging observations are used to establish a long-term behavioral representation, while a rolling window continuously captures recent charging dynamics and their deviations from this representation. The resulting changes are semantically encoded to guide drift-aware model adaptation, followed by quantile-based forecasting and online calibration to produce reliable probabilistic predictions under evolving charging behaviors.

\subsection{Dual-Timescale Behavior Modeling}
EV charging behavior exhibits both persistent station-specific patterns and short-term temporal variations. The former characterize a station's typical charging activity and temporal regularity, whereas the latter reflect recent changes relative to these established patterns. To distinguish these complementary characteristics, we construct a dual-timescale behavior representation comprising a fixed long-term behavior anchor and a rolling recent behavior state. Specifically, a behavior operator $\boldsymbol{\phi}(\cdot)$ extracts seven load-derived descriptors from an observation window $\mathcal{W}$:
\begin{equation}
\boldsymbol{\phi}(\mathcal{W})
=
[\mu,\sigma,z,h^{\mathrm{pk}},r,p^{\mathrm{hi}},v^{\mathrm{day}}]^{\top},
\label{eq:behavior_operator}
\end{equation}
where $\mu$, $\sigma$, $z$, $h^{\mathrm{pk}}$, $r$, $p^{\mathrm{hi}}$, and $v^{\mathrm{day}}$ denote the mean charging activity, load volatility, zero-load ratio, peak-hour index, ramping intensity, high-load fraction, and day-to-day variability, respectively. Together, these descriptors characterize charging activity, inactivity, peak timing, and temporal variability, providing interpretable proxies for station-level charging behavior.

The long-term behavior anchor of station $s$ is computed from its initial $D_L$-day historical observations as
$\mathbf{b}_{s}^{L}
=
\boldsymbol{\phi}(\mathcal{H}_{s}^{D_L})$
and remains fixed during online forecasting, providing a consistent station-specific reference. In contrast, the recent behavior state is updated using the latest $D_R$ hourly observations:
\begin{equation}
\mathbf{b}_{s,t}^{R}
=
\boldsymbol{\phi}
\left(
\{y_{s,\tau}\}_{\tau=t-D_R+1}^{t}
\right),
\label{eq:recent_behavior}
\end{equation}
where $D_R=168$ corresponds to the seven-day rolling window used in this study. Comparing the recent state with the long-term anchor identifies changes in charging activity, volatility, inactivity, peak timing, and other behavioral characteristics. These descriptor-level states and deviations subsequently provide the basis for constructing semantic prompts that describe the station's persistent operating pattern and its recent behavioral evolution.

Meanwhile, the historical load sequence and target-time calendar information are encoded to capture the temporal dynamics underlying charging demand. Specifically, the normalized historical input $\widetilde{\mathbf{x}}_{s,t}$ and calendar covariates $\mathbf{c}_{s,t+H}$ are mapped into a temporal representation as
\begin{equation}
\mathbf{h}_{s,t}^{\mathrm{TS}}
=
\mathcal{F}\left(
[\widetilde{\mathbf{x}}_{s,t};
\mathbf{c}_{s,t+H}]
\right)
\label{eq:temporal_representation}
\end{equation}
where $\mathcal{F}$ denotes the temporal encoder. 

\subsection{Behavior-Semantic Prompt Encoding}
\begin{figure}
    \centering
    \includegraphics[width=0.99\linewidth]{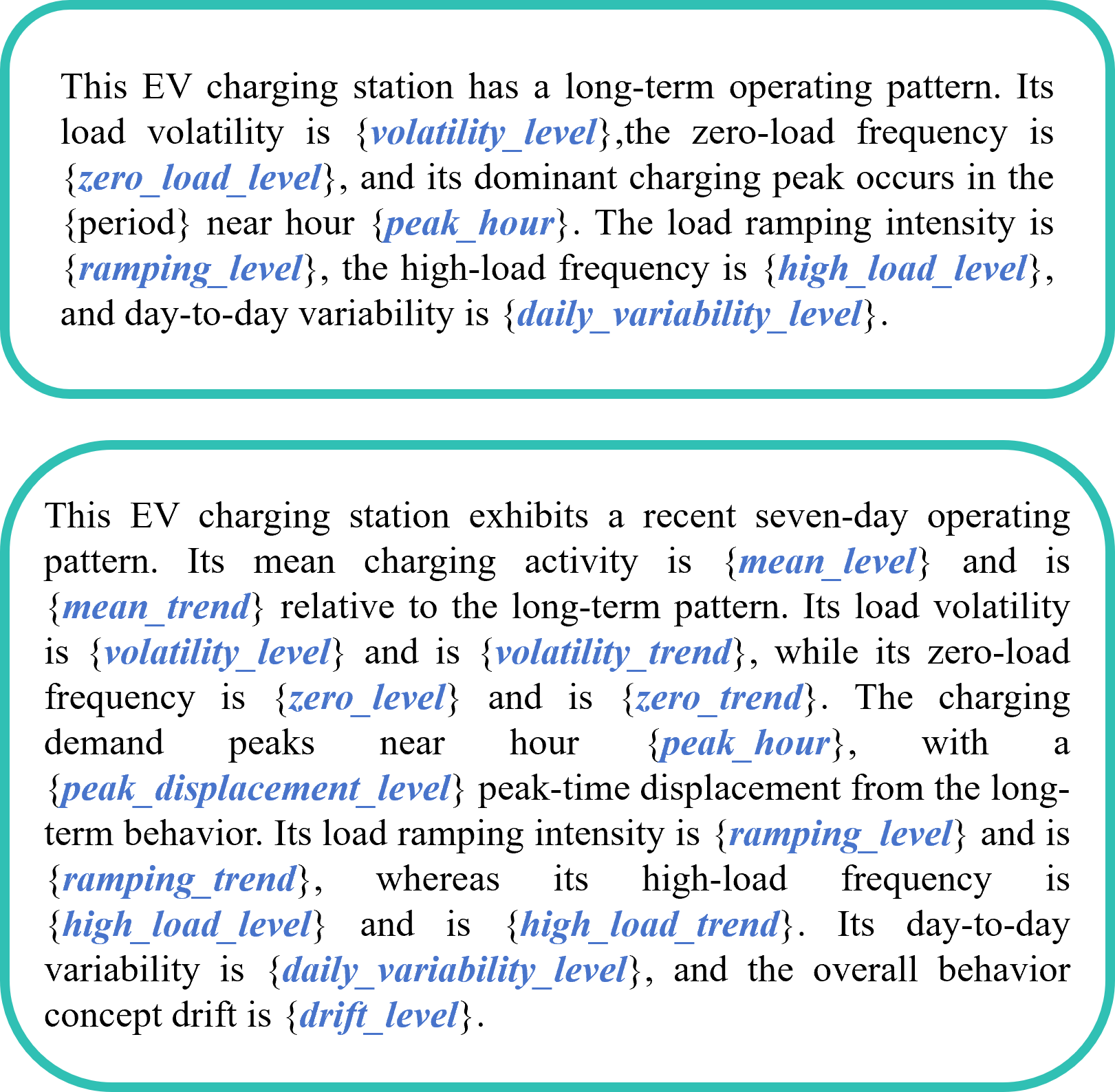}
    \caption{Semantic prompts for long-term and recent charging behaviors.}
    \label{fig:prompt}
\end{figure}

Although the numerical behavior descriptors quantify recent charging changes, they provide limited semantic interpretation of evolving behavior patterns. To leverage the semantic understanding capability of language models, we construct a descriptor-specific prompt for each behavioral characteristic based on its long-term and recent states:
\begin{equation}
\mathcal{P}_{s,t}
=
\left\{
P_j
\left(
b_{s,j}^{L},
b_{s,t,j}^{R},
\Delta b_{s,t,j}
\right)
\right\}_{j=1}^{J}
\label{eq:behavior_prompts}
\end{equation}
where $J$ denotes the number of behavior descriptors and $P_j(\cdot)$ describes the recent state and its change relative to the corresponding long-term behavior(see Figure~\ref{fig:prompt}).

Behavior prompt is encoded by a frozen language model $E_{\mathrm{LM}}(\cdot)$ and projected into a compact semantic space:
\begin{equation}
\widetilde{\mathbf{z}}_{s,t}
=
\left[\mathbf{W}E_{\mathrm{LM}}\left(P\right)
\right]
\label{eq:semantic_encoding}
\end{equation}
where $\mathbf{W}$ is a shared trainable projection matrix. The language model remains frozen throughout forecasting and adaptation and the resulting $\widetilde{\mathbf{z}}_{s,t}$ characterizes behavior changes and is subsequently used for behavior-guided adaptation.

\subsection{Behavior-Guided Online Probabilistic Forecasting}
The behavior representation, its deviation from the long-term pattern, and the semantic features are integrated to construct a behavior-aware residual correction. Its contribution is further controlled by the degree of behavior drift, allowing the model to adaptively respond to evolving charging patterns. The overall process is formulated as
\begin{subequations}
\label{eq:behavior_adaptation}
\begin{align}
\mathbf{u}_{s,t}
&=
\left[
\mathbf{b}_{s,t}^{R};
\Delta\mathbf{b}_{s,t};
\widetilde{\mathbf{z}}_{s,t}
\right]
\\
\mathbf{r}_{s,t}
&=
\mathbf{W}_{A,2}
\operatorname{GELU}
\left(
\mathbf{W}_{A,1}\mathbf{u}_{s,t}
\right)
\\
d_{s,t}
&=
\left\|
\mathbf{b}_{s,t}^{R}
-
\mathbf{b}_{s}^{L}
\right\|_{2}
\\
g_{s,t}
&=
\operatorname{sigmoid}
\left(
\frac{d_{s,t}-\mu_{d,t}}
{\sigma_{d,t}+\epsilon}
\right)
\\
\eta_{s,t}
&=
\eta_{0}
\left[
\lambda_{\min}
+
\left(
\lambda_{\max}-\lambda_{\min}
\right)g_{s,t}
\right]
\end{align}
\end{subequations}
where $\mathbf{u}_{s,t}$ denotes the integrated behavior representation, $\mathbf{W}_{A,1}$ and $\mathbf{W}_{A,2}$ are the trainable adapter matrices, and $\mathbf{r}_{s,t}$ is the resulting behavior-aware residual. The drift magnitude $d_{s,t}$ measures the deviation of recent behavior from the long-term anchor, while $\mu_{d,t}$ and $\sigma_{d,t}$ denote the historical mean and standard deviation of the drift, respectively, and $\epsilon$ is a small constant for numerical stability. The resulting $g_{s,t}\in(0,1)$ represents the drift intensity and jointly controls the residual contribution and online learning rate $\eta_{s,t}$, where $\eta_{0}$ is the base learning rate and $\lambda_{\min}$ and $\lambda_{\max}$ specify its lower and upper scaling factors.

The behavior-aware residual is then incorporated into the temporal representation:
\begin{equation}
\mathbf{h}_{s,t}
=
\mathbf{h}_{s,t}^{\mathrm{TS}}
+
g_{s,t}\mathbf{r}_{s,t}
\label{eq:representation_fusion}
\end{equation}
where $\mathbf{h}_{s,t}$ represents the behavior-adapted forecasting representation. In this manner, the temporal encoder provides the primary forecasting representation, while the behavior branch introduces an adaptive correction according to the observed charging-pattern changes.

Finally, the adapted representation $\mathbf{h}_{s,t}$ is mapped to the probabilistic forecasting output through a monotonic quantile decoder:
\begin{subequations}
\label{eq:quantile_decoder}
\begin{align}
&[a_{s,t},w_{s,t}^{-},w_{s,t}^{+}]=
F_Q\left(\mathbf{h}_{s,t}\right)
\\
&\widehat{q}_{s,t+H}^{0.50}=
a_{s,t}
\\
&\widehat{q}_{s,t+H}^{0.05}=
a_{s,t}-\operatorname{softplus}(w_{s,t}^{-})
\\
&\widehat{q}_{s,t+H}^{0.95}=
a_{s,t}+\operatorname{softplus}(w_{s,t}^{+})
\end{align}
\end{subequations}
where the resulting quantiles satisfy
$\widehat{q}_{s,t+H}^{0.05}\leq
\widehat{q}_{s,t+H}^{0.50}\leq
\widehat{q}_{s,t+H}^{0.95}$ by construction, providing the median forecast and the corresponding $90\%$ prediction interval.The complete algorithm process can be seen in Algorithm~\ref{alg:online_forecasting}.
\begin{figure*}[!t]
    \centering
    \includegraphics[width=0.99\linewidth]{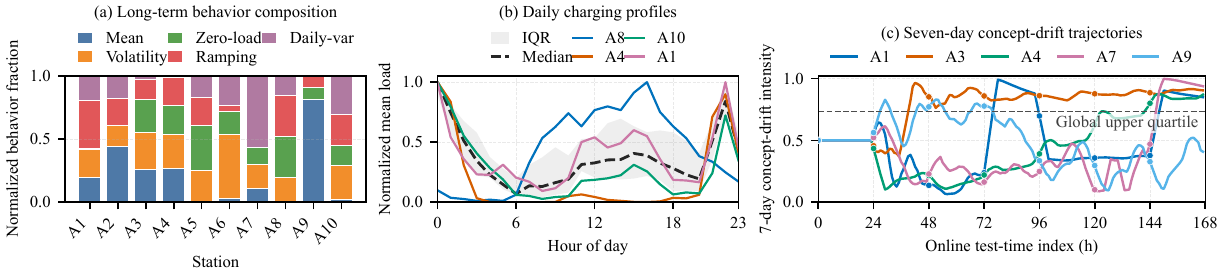}
    \caption{Heterogeneous and evolving EV charging behaviors across stations.}
    \label{fig:dataset}
\end{figure*}

\begin{algorithm}[t]
\caption{Behavior-Guided Online Probabilistic Forecasting}
\label{alg:online_forecasting}
\begin{algorithmic}[1]

\Require Charging stream $\mathcal{H}_{s,t}$, horizon $H$,
look-back $L$, windows $D_L=60$ days and $D_R$,
model $f_{\boldsymbol{\theta}}$
\Ensure Probabilistic forecasts $\widehat{\mathbf{q}}_{s,t+H}$

\State Initialize long-term behavior
$\mathbf{b}_{s}^{L}
=\boldsymbol{\phi}(\mathcal{H}_{s}^{D_L})$

\State Pretrain $f_{\boldsymbol{\theta}}$ using Pinball loss
on samples whose inputs and targets lie entirely within
the first $60$ days; keep the language model frozen

\For{each forecast origin $t$ after the initial $60$ days}

    \State Encode temporal representation
    $\mathbf{h}_{s,t}^{\mathrm{TS}}
    =F_{\mathrm{TS}}(\widetilde{\mathbf{x}}_{s,t},
    \mathbf{c}_{s,t+H})$

    \State Extract recent behavior $\mathbf{b}_{s,t}^{R}$ and deviation
    $\Delta\mathbf{b}_{s,t}
    =\mathbf{b}_{s,t}^{R}-\mathbf{b}_{s}^{L}$

    \State Encode behavior prompts into semantic feature
    $\widetilde{\mathbf{z}}_{s,t}$

    \State Compute residual $\mathbf{r}_{s,t}$, drift intensity
    $g_{s,t}$, and learning rate $\eta_{s,t}$

    \State Adapt temporal representation
    $\mathbf{h}_{s,t}
    =\mathbf{h}_{s,t}^{\mathrm{TS}}
    +g_{s,t}\mathbf{r}_{s,t}$

    \State Generate quantile forecast
    $\widehat{\mathbf{q}}_{s,t+H}
    =F_Q(\mathbf{h}_{s,t})$

    \State Store pending sample
    $\pi_{s,t}^{(H)}
    =(s,t,t+H,\mathbf{x}_{s,t},
    \mathbf{c}_{s,t+H},
    \widehat{\mathbf{q}}_{s,t+H})$

    \State Retrieve matured set $\mathcal{M}_{t}^{(H)}$

    \If{$\mathcal{M}_{t}^{(H)}\neq\emptyset$}

        \State Evaluate gradients using matured observations
        $\mathbf{g}_{s,\tau\rightarrow t}^{(H)}
        =
        \left.
        \nabla_{\boldsymbol{\theta}}
        \mathcal{L}_{\mathrm{PB}}
        \right|_{\boldsymbol{\theta}
        =\boldsymbol{\theta}_{t}}$

        \State Update current model parameters
        $\boldsymbol{\theta}_{t^{+}}
        =
        \boldsymbol{\theta}_{t}
        -
        \sum_{s\in\mathcal{S}}\eta_{s,t}
        \sum_{\pi_{s,\tau}^{(H)}
        \in\mathcal{M}_{t}^{(H)}}
        \mathbf{g}_{s,\tau\rightarrow t}^{(H)}$

    \EndIf

\EndFor

\end{algorithmic}
\end{algorithm}

\section{Experiment}
\subsection{Setting}
In this section, we evaluate the proposed approach using the real-world multi-prototype EV charging dataset introduced in \cite{liu2026ai}, which contains ten heterogeneous charging stations with diverse charging behaviors and operating characteristics. The proposed method is compared against five representative baselines: TCN\cite{tian2025short}, PatchTST\cite{nie2023patchtst}, LSTM, FSNet\cite{pham2023fsnet}, and OneNet\cite{zhang2023onenet}(Details in Appendix). Since this study focuses on online forecasting, the first 60 days of observations are used as the initial historical period, after which the remaining data are processed sequentially in a rolling online manner. At each forecast origin, probabilistic forecasts are generated using only the information available at that time, and model updates are performed only when the corresponding observations become available. Both 1-h-ahead and 4-h-ahead forecasting tasks are considered. Deep learning models are implemented in PyTorch with CUDA support (PyTorch version 2.4.1 on an NVIDIA GeForce RTX 3090 GPU), and all experiments are conducted on a workstation with an Intel Core i7-13900HX CPU and 16 GB RAM. The performance is assessed using five metrics: MSE, Pinball loss, PICP@90, PINAW@90, and CD@90(Details in Appendix). LSTM is used as the temporal encoder.

\subsection{Dataset}
Fig.~\ref{fig:dataset} provides an overview of the heterogeneous and evolving charging behaviors across the ten selected stations. As shown in Fig.~\ref{fig:dataset}(a), the stations exhibit distinct long-term behavior compositions in terms of mean activity, volatility, zero-load ratio, ramping intensity, and day-to-day variability. Fig.~\ref{fig:dataset}(b) further shows substantial differences in their daily charging profiles, including the timing and magnitude of charging peaks as well as within-day variability. Beyond these persistent station-specific patterns, Fig.~\ref{fig:dataset}(c) reveals that the behavioral states continue to evolve during the online forecasting period. In particular, the seven-day drift trajectories vary considerably across stations and over time, with several stations exhibiting pronounced deviations from their historical behavior. These observations demonstrate that EV charging loads are characterized by both cross-station behavioral heterogeneity and time-varying behavioral drift, motivating the proposed behavior modeling and online adaptation.

\begin{figure*}[!t]
    \centering
    \includegraphics[width=0.99\linewidth]{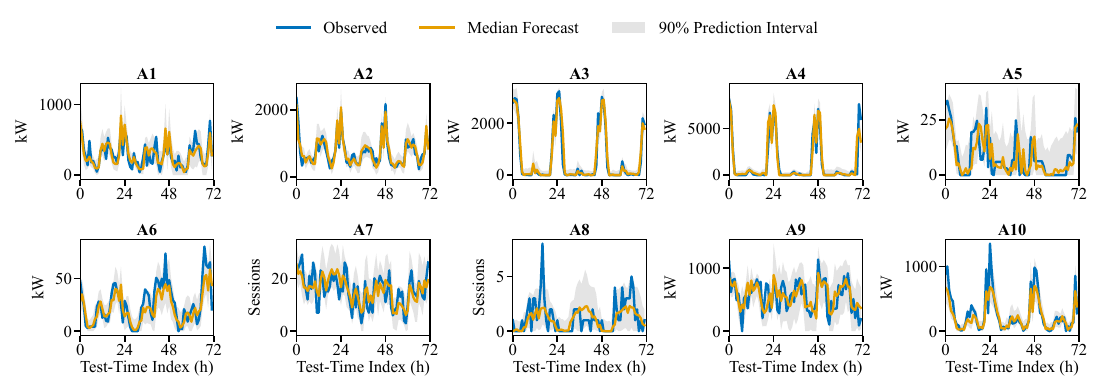}
    \caption{Online probabilistic forecasts across heterogeneous EV charging stations.}
    \label{fig:visual}
\end{figure*}

\subsection{Result Analysis}
Fig.~\ref{fig:visual} presents the online probabilistic forecasting results across the ten heterogeneous charging stations, showing that the proposed method can effectively track the observed charging dynamics despite substantial differences in load magnitude, peak characteristics, 
and temporal patterns. The quantitative results in Table~\ref{tab:forecasting_comparison} further compare the forecasting performance over the 1-h and 4-h horizons. Overall, TCN, PatchTST, and LSTM exhibit relatively limited performance under the online setting. For the 1-h-ahead task, their MSE values are 0.653, 0.678, and 0.674, respectively, while the corresponding PICP@90 values are only 0.751, 0.812, and 0.751. This indicates that conventional temporal representations combined with online updates remain insufficient to simultaneously accommodate evolving charging patterns and maintain reliable probabilistic coverage. Although PatchTST improves PICP@90 to 0.812, its PINAW@90 increases to 0.357, indicating that the improved coverage is accompanied by relatively wider prediction intervals. In comparison, FSNet and OneNet, which are specifically designed to address non-stationarity and concept drift in online time-series forecasting, achieve substantially better overall performance. Their 1-h Pinball losses decrease to 0.124 and 0.118, respectively, while PICP@90 improves to 0.848 and 0.873, demonstrating the importance of adapting forecasting models to evolving temporal distributions.

The proposed method further improves upon these strong online baselines. For the 1-h-ahead task, it achieves an MSE of 0.360 and a Pinball loss of 0.097, representing reductions of 15.3\% and 17.8\% over the corresponding best baseline results of FSNet (0.425) and OneNet (0.118), respectively. More importantly, the improved coverage is not achieved by simply widening the prediction intervals. The proposed method obtains 
a PICP@90 of 0.897, close to the nominal coverage level of 0.900, while reducing PINAW@90 to 0.201, compared with 0.228 for OneNet and 0.240 for FSNet, resulting in a CD@90 of only 0.002. This demonstrates a better balance between probabilistic reliability and interval sharpness. When the forecasting horizon increases from 1 h to 4 h, all baseline methods exhibit varying degrees of performance degradation. In particular, the 
MSE of TCN increases from 0.653 to 0.984, while its PICP@90 decreases from 0.751 to 0.703; the coverage levels of PatchTST and LSTM also decrease to 0.768 and 0.743, respectively. FSNet and OneNet show greater robustness to the increased forecasting horizon, but their PICP@90 values remain at 0.852 and 0.844. In contrast, the proposed method maintains an MSE of 0.412, a Pinball loss of 0.106, a PINAW@90 of 0.223, and a CD@90 of 0.003, while preserving a PICP@90 of 0.897. These results indicate that, compared with conventional online updating and general concept-drift adaptation, explicitly characterizing persistent station-specific behaviors and their recent deviations provides more informative guidance for online adaptation, thereby maintaining forecasting accuracy, interval sharpness, and probabilistic calibration under evolving charging behaviors and extended forecasting horizons.
\begin{table*}[!t]
    \centering
    \caption{Performance comparison over 1 h and 4 h forecasting horizons.
    The best result is highlighted in bold red, while the second-best
    result is underlined in blue.}
    \label{tab:forecasting_comparison}

    \begin{adjustbox}{max width=\textwidth}
    \begin{tabular}{lccccc@{\hspace{6pt}}ccccc}
        \toprule
        & \multicolumn{5}{c}{\textbf{1 h ahead}}
        & \multicolumn{5}{c}{\textbf{4 h ahead}} \\
        \cmidrule(lr){2-6}
        \cmidrule(lr){7-11}

        \textbf{Model}
        & \textbf{MSE$\downarrow$}
        & \textbf{Pinball$\downarrow$}
        & \textbf{PICP@90}
        & \textbf{PINAW@90$\downarrow$}
        & \textbf{CD@90$\downarrow$}
        & \textbf{MSE$\downarrow$}
        & \textbf{Pinball$\downarrow$}
        & \textbf{PICP@90}
        & \textbf{PINAW@90$\downarrow$}
        & \textbf{CD@90$\downarrow$} \\
        \midrule

        TCN
        & 0.653 & 0.173 & 0.751 & 0.329 & 0.149
        & 0.984 & 0.239 & 0.703 & 0.430 & 0.197 \\




        PatchTST
        & 0.678 & 0.179 & 0.812 & 0.357 & 0.088
        & 0.923 & 0.212 & 0.768 & 0.417 & 0.132 \\

        LSTM
        & 0.674 & 0.164 & 0.751 & 0.362 & 0.149
        & 0.896 & 0.225 & 0.743 & 0.391 & 0.157 \\

        FSNet
        & \second{0.425}
        & 0.124
        & 0.848
        & 0.240
        & {0.052}
        & {0.516}
        & {0.142}
        & \second{0.852}
        & 0.283
        & 0.048 \\

        OneNet
        & {0.453}
        & \second{0.118}
        & \second{0.873}
        & \second{0.228}
        & \second{0.027}
        & \second{0.495}
        & \second{0.137}
        & 0.844
        & \second{0.278}
        & \second{0.056} \\

        Proposed
        & \best{0.360}
        & \best{0.097}
        & \best{0.898}
        & \best{0.201}
        & \best{0.002}
        & \best{0.412}
        & \best{0.106}
        & \best{0.897}
        & \best{0.223}
        & \best{0.003} \\

        \bottomrule
    \end{tabular}
    \end{adjustbox}
    \begin{minipage}{\textwidth}
        \footnotesize
        \textit{Note:} Lower values are better for MSE, Pinball,
        PINAW@90, and CD@90. For PICP@90, values closer to the nominal coverage level of 0.900 are better.
    \end{minipage}
\end{table*}

\subsection{Ablation Study}
\begin{figure}
    \centering
    \includegraphics[width=0.99\linewidth]{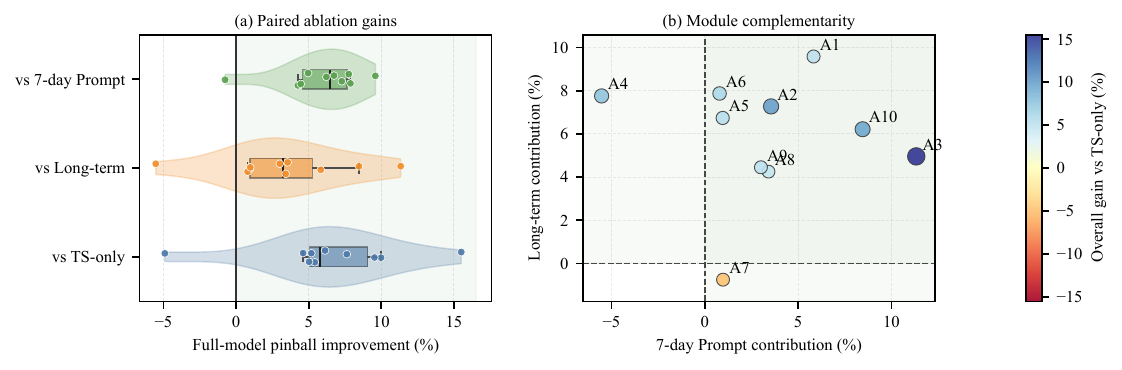}
    \caption{Ablation analysis of behavior-guided forecasting components.}
    \label{fig:ab1}
\end{figure}

\begin{figure}
    \centering
    \includegraphics[width=0.99\linewidth]{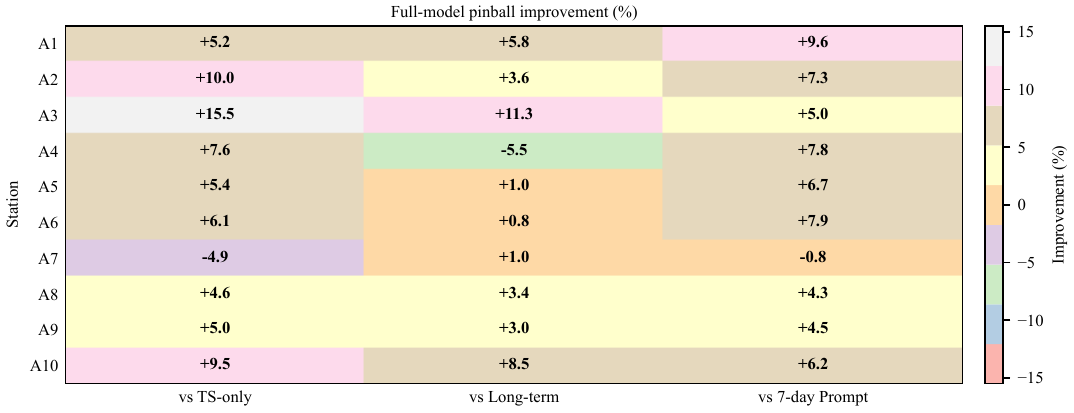}
    \caption{Station-wise ablation analysis of behavior-guided components.}
    \label{fig:ab2}
\end{figure}

Figs.~\ref{fig:ab1} and~\ref{fig:ab2} further investigate the contributions of the proposed behavior-guided components. As shown in Fig.~\ref{fig:ab1}(a), the full model generally outperforms the TS-only, long-term, and 7-day-prompt variants, confirming the effectiveness of incorporating behavioral information into online probabilistic forecasting. Fig.~\ref{fig:ab1}(b) further reveals a clear complementarity between the two behavior components, as most stations benefit from jointly incorporating long-term behavioral patterns and recent behavioral changes. The station-wise results in Fig.~\ref{fig:ab2} show that the magnitude of these improvements varies considerably across heterogeneous charging stations. In particular, some stations benefit more from the long-term representation, whereas others exhibit stronger dependence on recent behavioral information, reflecting differences in their operating regularity and temporal variability. A few stations show limited or negative gains from individual components, indicating that behavioral information is not equally informative under all operating patterns. Nevertheless, the combined framework provides consistent improvements for the majority of stations.

\begin{figure}
    \centering
    \includegraphics[width=0.99\linewidth]{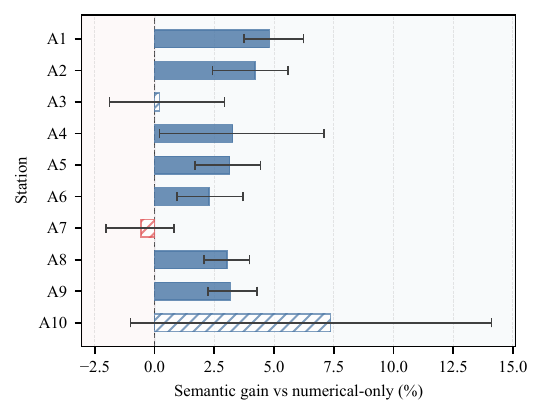}
    \caption{Station-wise gains of semantic over numerical behavior encoding.}
    \label{fig:semantic_gain}
\end{figure}

\begin{figure}
    \centering
    \includegraphics[width=0.99\linewidth]{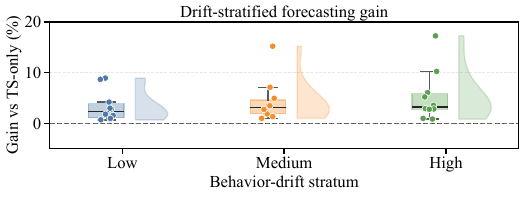}
    \caption{Forecasting gains across different behavior-drift intensities.}
    \label{fig:drift_gain}
\end{figure}

To further validate the effectiveness of behavior-guided adaptation and semantic encoding, additional experiments are conducted. As shown in Fig.~\ref{fig:drift_gain}, forecasting gains generally increase with behavior-drift intensity, indicating that behavioral information becomes more valuable as recent charging patterns deviate from persistent behaviors. The larger gains under high-drift conditions further demonstrate that the proposed adaptation mechanism can effectively respond to pronounced behavioral changes rather than providing uniform corrections. Fig.~\ref{fig:semantic_gain} shows that semantic encoding outperforms direct numerical encoding at most stations, demonstrating its ability to extract additional information from behavioral descriptors. The station-dependent gains also suggest that semantic information is particularly beneficial when charging behaviors exhibit complex temporal variations that cannot be fully represented by numerical descriptors alone. Overall, these results confirm that explicitly characterizing behavioral evolution and semantically interpreting such changes provide complementary guidance for online forecasting adaptation.

\subsection{Efficiency Analysis}
Table~\ref{tab:computational_efficiency} evaluates the computational efficiency of different forecasting methods. Although the proposed method contains substantially more parameters due to the behavior-semantic module, its inference and online update times remain at 1.85 ms/sample and 8.67 ms/step, respectively, demonstrating practical computational 
efficiency for online forecasting. Its inference latency is lower than that of PatchTST and remains comparable to other online baselines, while the P95 latency is maintained at 20.62 ms. The memory consumption of 17.4 MB also remains moderate under the experimental platform. These results indicate that the additional behavior-guided modeling introduces manageable computational overhead while preserving efficient online forecasting and adaptation.

\begin{table}[htbp] 
    \centering 
    \caption{Computational efficiency and resource consumption.} 
    \label{tab:computational_efficiency} 
    \small 
    \resizebox{\linewidth}{!}{%
    \begin{tabular}{lrrrrr} 
        \toprule 
        \textbf{Model} 
        & \textbf{Params.} 
        & \textbf{Inference} 
        & \textbf{Update} 
        & \textbf{P95} 
        & \textbf{Memory} \\ 
        & \textbf{(M)} 
        & \textbf{(ms/sample)} 
        & \textbf{(ms/step)} 
        & \textbf{(ms)} 
        & \textbf{(MB)} \\ 
        \midrule 
 
        TCN 
        & 0.06 & 0.61 & 3.48 & 4.58 & 18.2 \\ 
 
        PatchTST 
        & 0.60 & 2.15 & 9.24 & 14.79 & 11.4 \\ 
 
        LSTM 
        & 0.05 & 0.21 & 2.97 & 4.21 & 1.8 \\ 
 
        FSNet 
        & 0.07 & 0.84 & 4.51 & 6.70 & 2.1 \\ 
 
        OneNet 
        & 0.13 & 1.42 & 6.99 & 11.30 & 3.0 \\ 
 
        Proposed 
        & 81.93 & 1.85 & 8.67 & 20.62 & 17.4 \\ 
 
        \bottomrule 
    \end{tabular}%
    } 
\end{table}

\subsection{Scalability Analysis}
To assess the generality of the proposed framework, we replace the default LSTM encoder with TCN and PatchTST, representing convolutional and patch-based attention architectures, respectively. Each encoder is evaluated with and without behavior-guided adaptation, while maintaining identical initial 60-day training periods, forecasting horizons, quantile levels, and delayed-feedback protocols. The behavior-guided variants share the same dual-timescale representation, frozen semantic encoder, residual adapter, and drift-aware learning-rate mechanism. Table~\ref{tab:encoder_generality} compares MSE, Pinball loss, and PICP@90 to assess point accuracy, probabilistic accuracy, and interval coverage. The behavior-guided TCN and PatchTST results are currently illustrative placeholders awaiting experimental verification. This comparison will establish whether behavioral guidance improves forecasting accuracy and brings coverage closer to the nominal level of 0.90 across different encoders. Consistent improvements would support the framework's applicability beyond recurrent architectures, while variations in improvement magnitude would reveal how behavioral information complements different temporal representations.

\begin{table}[t]
\centering
\caption{Evaluating the scalability of behavior-guided forecasting.}
\label{tab:encoder_generality}
\renewcommand{\arraystretch}{1.15}
\setlength{\tabcolsep}{3pt}
\resizebox{\columnwidth}{!}{%
\begin{tabular}{llcccccc}
\toprule
\multirow{2}{*}{Encoder}
& \multirow{2}{*}{Behavior}
& \multicolumn{3}{c}{1 h ahead}
& \multicolumn{3}{c}{4 h ahead} \\
\cmidrule(lr){3-5}
\cmidrule(lr){6-8}
& & MSE$\downarrow$ & Pinball$\downarrow$ & PICP@90
  & MSE$\downarrow$ & Pinball$\downarrow$ & PICP@90 \\
\midrule

LSTM & Without
& 0.674 & 0.164 & 0.751
& 0.896 & 0.225 & 0.743 \\
\rowcolor{gray!15}
LSTM & With
& 0.360 & 0.097 & 0.898
& 0.412 & 0.106 & 0.897 \\

\midrule
TCN & Without
& 0.653 & 0.173 & 0.751
& 0.984 & 0.239 & 0.703 \\
\rowcolor{gray!15}
TCN & With
& 0.378 & 0.103 & 0.891
& 0.446 & 0.116 & 0.886 \\

\midrule
PatchTST & Without
& 0.678 & 0.179 & 0.812
& 0.923 & 0.212 & 0.768 \\
\rowcolor{gray!15}
PatchTST & With
& 0.365 & 0.104 & 0.895
& 0.429 & 0.127 & 0.892 \\

\bottomrule
\end{tabular}%
}
\end{table}

\section{Conclusion}
This paper proposed a behavior-guided online probabilistic forecasting framework for EV charging loads under heterogeneous and evolving charging behaviors. The framework integrates persistent station-specific patterns with recent behavioral changes through dual-timescale behavior modeling and semantic encoding, enabling behavior-aware adaptation as new observations become available. Experiments on ten heterogeneous charging stations demonstrated that the proposed method consistently outperformed conventional forecasting models and concept-drift-aware online baselines in both 1-h and 4-h forecasting. The results further showed improved probabilistic accuracy, interval sharpness, and calibration, while the ablation analysis confirmed the complementary roles of long-term and recent behavioral information. Despite the additional behavior modeling, the computational overhead remained manageable for online implementation. Overall, the results highlight the value of explicitly incorporating evolving charging behaviors into online EV load forecasting. Future work will incorporate external behavioral factors and investigate the generalizability of the framework across broader charging environments.

\section*{Appendix}
\subsection{Metric}
The forecasting performance is evaluated from the perspectives of point accuracy, probabilistic accuracy, interval coverage, and calibration. Let $y_i$ denote the observed charging load, 
$\widehat{y}_i$ the median forecast, and $\widehat{q}_i^{\tau}$ the predicted quantile at quantile level $\tau\in\mathcal{Q}$, where $N$ denotes the total number of online forecasting samples.

The point forecasting accuracy is evaluated using the MSE:
\begin{equation}
    \mathrm{MSE}
    =
    \frac{1}{N}
    \sum_{i=1}^{N}
    \left(
        y_i-\widehat{y}_i
    \right)^2
    \label{eq:mse_metric}
\end{equation}

The probabilistic forecasting accuracy is evaluated using the average
pinball loss:
\begin{equation}
    \mathrm{Pinball}
    =
    \frac{1}{N|\mathcal{Q}|}
    \sum_{i=1}^{N}
    \sum_{\tau\in\mathcal{Q}}
    \rho_{\tau}
    \left(
        y_i-\widehat{q}_i^{\tau}
    \right)
    \label{eq:pinball_metric}
\end{equation}
where $\rho_{\tau}(e)=\max\{\tau e,(\tau-1)e\}$ denotes the pinball loss at quantile level $\tau$, with a lower value indicating better probabilistic forecasting performance.

For the nominal $90\%$ prediction interval
$\widehat{I}_{i}^{0.90}=[\widehat{q}_{i}^{0.05},\widehat{q}_{i}^{0.95}]$,
the prediction interval coverage probability (PICP@90) is defined as
\begin{equation}
    \mathrm{PICP@90}
    =
    \frac{1}{N}
    \sum_{i=1}^{N}
    \mathbb{I}
    \left\{
        \widehat{q}_{i}^{0.05}
        \leq y_i
        \leq
        \widehat{q}_{i}^{0.95}
    \right\}
    \label{eq:picp_metric}
\end{equation}
where $\mathbb{I}\{\cdot\}$ denotes the indicator function. A
well-calibrated $90\%$ prediction interval should yield a PICP close to $0.90$.

The sharpness of the prediction interval is measured by the prediction interval normalized average width (PINAW@90):
\begin{equation}
    \mathrm{PINAW@90}
    =
    \frac{1}{N R_y}
    \sum_{i=1}^{N}
    \left(
        \widehat{q}_{i}^{0.95}
        -
        \widehat{q}_{i}^{0.05}
    \right)
    \label{eq:pinaw_metric}
\end{equation}
where $R_y=y_{\max}-y_{\min}$ denotes the range of the observed charging load over the evaluation set, with a lower PINAW indicating a sharper prediction interval.

To explicitly quantify the deviation between empirical and nominal coverage, the coverage deviation (CD@90) is defined as
\begin{equation}
    \mathrm{CD@90}
    =
    \left|
        \mathrm{PICP@90}-0.90
    \right|.
    \label{eq:cd_metric}
\end{equation}

\subsection{Setting Details}
The detailed training and online adaptation settings are summarized in Table \ref{tab:training_online_settings}.
\begin{table}[!htbp]
\centering
\caption{Training and Online Adaptation Settings}
\label{tab:training_online_settings}
\footnotesize
\setlength{\tabcolsep}{3.5pt}
\renewcommand{\arraystretch}{1.06}
\begin{tabular}{@{}p{0.60\columnwidth}p{0.30\columnwidth}@{}}
\toprule
\textbf{Name} & \textbf{Setting} \\
\midrule

\multicolumn{2}{@{}l}{\textit{Offline pretraining}} \\
Initial training period & 60 days \\
Pretraining epochs & 30 \\
Batch size & 128 \\
Optimizer & AdamW \\
Initial learning rate & $3\times10^{-4}$ \\
Weight decay & $1\times10^{-5}$ \\
Quantile levels & $\{0.05,0.50,0.95\}$ \\
Training objective & Pinball loss \\
Gradient clipping norm & 1.0 \\
Random seed & 42 \\

\midrule
\multicolumn{2}{@{}l}{\textit{Online adaptation}} \\
Recent behavior window & 7 days (168 h) \\
Online batch size & 1 \\
Online updates per arrived label & 1 \\
Online optimizer & AdamW \\
Base online learning rate & $3\times10^{-4}$ \\
Online weight decay & $1\times10^{-5}$ \\
Learning-rate factor range & $[0.25,\,2.0]$ \\
Effective learning-rate range
& $[7.5\times10^{-5},\,6\times10^{-4}]$ \\
Drift-history window & 168 samples \\
Drift warm-up period & 24 samples \\
Initial drift intensity & 0.5 \\
Online objective & Pinball loss \\
Gradient clipping norm & 1.0 \\
Label delay & $H\in\{1,4\}$ h \\
Frozen semantic encoder & DistilGPT2 \\
\bottomrule
\end{tabular}
\end{table}

\subsection{Baselines}
\textit{TCN}: TCN employs temporal convolutional operations to capture local and long-range dependencies in charging load sequences. It is implemented with online updates to provide a convolution-based baseline under the same forecasting setting.

\textit{PatchTST}: PatchTST divides time-series observations into patches and employs a Transformer architecture to model temporal dependencies. It serves as a representative Transformer-based forecasting baseline.

\textit{LSTM}: LSTM captures sequential dependencies through recurrent hidden states and gated memory mechanisms. It is adopted as a representative recurrent baseline for online EV charging load forecasting.

\textit{FSNet}: FSNet combines fast and slow learning mechanisms to adapt forecasting models to evolving time-series distributions. It is included as a dedicated online forecasting baseline for handling temporal distribution shifts.

\textit{OneNet}: OneNet employs online ensembling to dynamically combine forecasting models under concept drift. It provides a strong baseline for evaluating adaptation to non-stationary charging load patterns.

\bibliographystyle{IEEEtran}
\bibliography{ref}
\end{document}